\documentclass[11pt]{article}

\usepackage[final]{acl}

\usepackage{times}
\usepackage{latexsym}

\usepackage[T1]{fontenc}

\usepackage[utf8]{inputenc}

\usepackage{microtype}

\usepackage{inconsolata}

\usepackage{graphicx}

\usepackage[switch]{lineno}
\usepackage{marvosym}
\usepackage{footmisc}
\usepackage{xurl}
\usepackage{xcolor}
\definecolor{my-purple}{RGB}{223,211,231}
\definecolor{my-yellow}{RGB}{255,242,204}
\definecolor{my-red}{RGB}{239,134,131}
\usepackage{enumitem}
\setlist[itemize]{leftmargin=*}
\usepackage{pfnote}
\usepackage{booktabs}
\usepackage{makecell}
\usepackage{multirow}
\usepackage[figuresleft]{rotating}
\usepackage{amsmath}

\title{Human-in-the-Loop Generation of Adversarial Texts:\\A Case Study on Tibetan Script}

\makeatletter
\def\thanks#1{\protected@xdef\@thanks{\@thanks\protect\footnotetext{#1}}}
\makeatother

\author{
	\textbf{Xi Cao$^{\heartsuit\spadesuit}$}\textbf{,}~
	\textbf{Yuan Sun$^{\heartsuit\spadesuit}$\textsuperscript{\Letter}}\textbf{,}~\\
	\textbf{Jiajun Li$^{\diamondsuit\clubsuit}$}\textbf{,}~
	\textbf{Quzong Gesang$^{\diamondsuit\clubsuit}$}\textbf{,}~
	\textbf{Nuo Qun$^{\diamondsuit\clubsuit}$\textsuperscript{\Letter}}\textbf{,}~
	\textbf{Tashi Nyima$^{\diamondsuit\clubsuit}$}\\
	\small{$^{\heartsuit}$Institute of National Security, Minzu University of China, Beijing, China}\\
	\small{$^{\diamondsuit}$School of Information Science and Technology, Xizang University, Lhasa, China}\\
	\small{$^{\spadesuit}$National Language Resource Monitoring \& Research Center | Minority Languages Branch, Beijing, China}\\
	\small{$^{\clubsuit}$The State Key Laboratory of Tibetan Intelligence, Lhasa, China}\\
	\normalsize{\texttt{caoxi@muc.edu.cn},
	\texttt{sunyuan@muc.edu.cn},
	\texttt{q\_nuo@utibet.edu.cn}}
	\thanks{\Letter~Corresponding Author}
}

\begin{document}
\maketitle
\begin{abstract}
DNN-based language models excel across various NLP tasks but remain highly vulnerable to textual adversarial attacks.
While adversarial text generation is crucial for NLP security, explainability, evaluation, and data augmentation, related work remains overwhelmingly English-centric, leaving the problem of constructing high-quality and sustainable adversarial robustness benchmarks for lower-resourced languages both difficult and understudied.
First, method customization for lower-resourced languages is complicated due to linguistic differences and limited resources.
Second, automated attacks are prone to generating invalid or ambiguous adversarial texts.
Last but not least, language models continuously evolve and may be immune to parts of previously generated adversarial texts.
To address these challenges, we introduce \texttt{HITL-GAT}\footnote{Video Demonstration:\\\url{https://youtu.be/tXla4yAggwA}\\Code Repository:\\\url{https://github.com/CMLI-NLP/HITL-GAT}\\Victim Models:\\\url{https://huggingface.co/collections/UTibetNLP/tibetan-victim-language-models-669f614ecea872c7211c121c}}, an interactive system based on a general approach to human-in-the-loop generation of adversarial texts.
Additionally, we demonstrate the utility of \texttt{HITL-GAT} through a case study on Tibetan script, employing three customized adversarial text generation methods and establishing its first adversarial robustness benchmark, providing a valuable reference for other lower-resourced languages.
\end{abstract}

\section{Introduction}

The adversarial attack refers to an attack method in which the attacker adds imperceptible perturbations to the original input, resulting in the incorrect judgment of a DNN-based model.
The examples generated during textual adversarial attacks are called adversarial texts.

\begin{figure}[t]
	\centering
	\includegraphics[width=\linewidth]{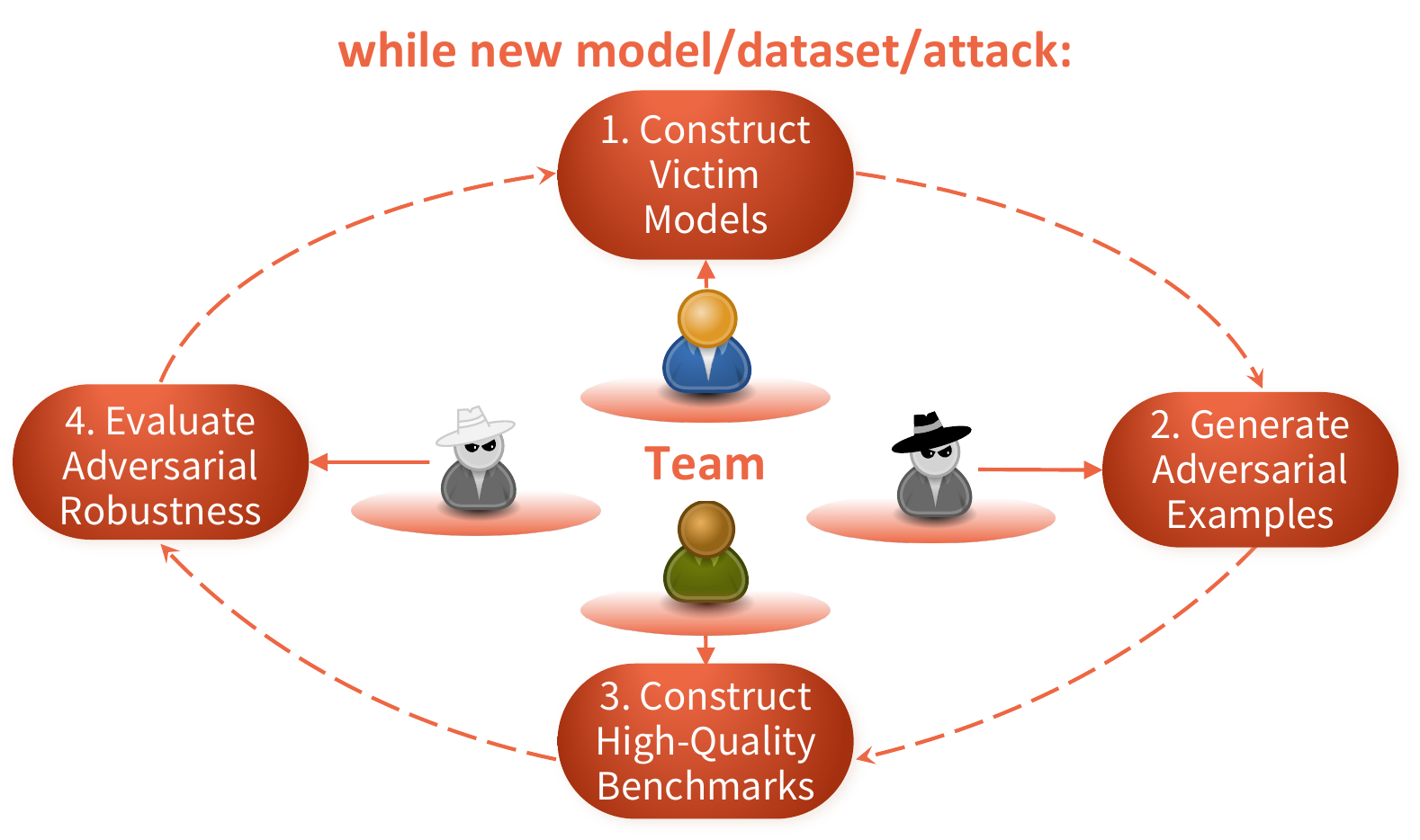}
	\captionsetup{justification=justified, singlelinecheck=false}
	\caption{Workflow of \texttt{HITL-GAT}. While a new language model, downstream dataset, or textual adversarial attack method emerges, we can enter the loop to make the adversarial robustness benchmark evolve.}
	\label{workflow}
\end{figure}

Due to the general adaptability of language models to classification tasks, adversarial robustness evaluation is mainly focused on the domain.
Currently, most of the adversarial text generation methods target higher-resourced languages, especially English.
Because of the differences in textual features and language resources, it is challenging to transfer these methods to other languages.
\textbf{Problem 1: How do we generate adversarial texts for lower-resourced languages?}

\citet{NEURIPS_DATASETS_AND_BENCHMARKS2021_335f5352} apply 14 textual adversarial attack methods to GLUE tasks \citep{DBLP:conf/iclr/WangSMHLB19} to construct the widely used adversarial robustness benchmark AdvGLUE.
In their construction, they find that most textual adversarial attack methods are prone to generating invalid or ambiguous adversarial texts, with around 90\% either changing the original semantics or hindering the annotators’ unanimity.
In our case study on Tibetan script, we also come to the same conclusion.
\textbf{Problem 2: How do we construct high-quality adversarial robustness benchmarks?}

\citet{wang2023on} employ ANLI \citep{nie-etal-2020-adversarial} and AdvGLUE \citep{NEURIPS_DATASETS_AND_BENCHMARKS2021_335f5352} to assess the adversarial robustness of ChatGPT and several previous popular language models and find ChatGPT is the best.
However, both ANLI and AdvGLUE are constructed using fine-tuned BERT \citep{devlin-etal-2019-bert} and RoBERTa \citep{liu2019robertarobustlyoptimizedbert} as victim models.
Language models are evolving, while adversarial robustness benchmarks never.
We argue that new language models may be immune to part of previously generated adversarial texts.
Lower-resourced languages are at a very early stage of adversarial robustness evaluation compared to higher-resourced languages, and it is essential to envisage sustainable adversarial robustness evaluation in advance.
\textbf{Problem 3: How do we update adversarial robustness benchmarks?}

\begin{figure*}[t]
	\centering
	\includegraphics[width=\linewidth]{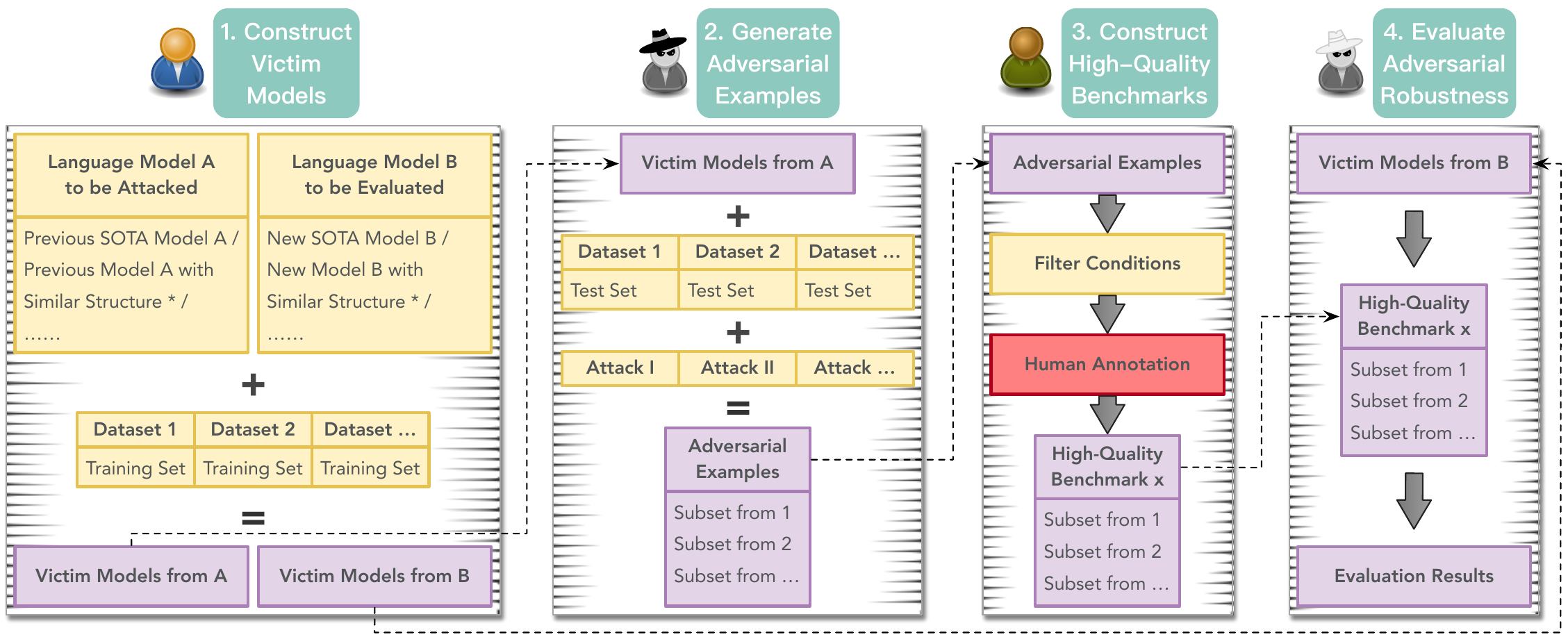}
	\captionsetup{justification=justified, singlelinecheck=false}
	\caption{Flowchart of \texttt{HITL-GAT}. Our system contains four stages in one pipeline: \textbf{victim model construction}, \textbf{adversarial example generation}, \textbf{high-quality benchmark construction}, and \textbf{adversarial robustness evaluation}. System outputs are highlighted in \colorbox{my-purple}{purple background}. Human choices are highlighted in \colorbox{my-yellow}{yellow background}. Human annotation is highlighted in \colorbox{my-red}{red background}.}
	\label{flowchart}
\end{figure*}

To address the above problems, we introduce \texttt{HITL-GAT}, an interactive system for human-in-the-loop generation of adversarial texts.
Figure \ref{workflow} depicts the workflow of \texttt{HITL-GAT}.
In a loop where a new language model, downstream dataset, or textual adversarial attack method emerges, our team starts to construct victim models, generate adversarial examples, construct high-quality benchmarks, and evaluate adversarial robustness.
The loop allows adversarial robustness benchmarks to evolve along with new models, datasets, and attacks \textbf{(Problem 3)}.
Figure \ref{flowchart} depicts the four stages in one pipeline detailedly.
Firstly, we fine-tune the previous model and the new model on the same downstream datasets to construct victim models.
Subsequently, we implement adversarial attacks on the victim models constructed from the previous model upon downstream datasets to generate adversarial examples.
Afterward, we customize filter conditions and conduct human annotation to construct a high-quality adversarial robustness benchmark \textbf{(Problem 2)}.
Finally, we evaluate the adversarial robustness of the new model on the benchmark.
Additionally, we make a case study on one lower-resourced language, Tibetan, based on the general human-in-the-loop approach to adversarial text generation \textbf{(Problem 1)}.

The contributions of this paper are as follows:

(1) We propose a general human-in-the-loop approach to adversarial text generation.
This approach can assist in constructing and updating high-quality adversarial robustness benchmarks with the emergence of new language models, downstream datasets, and textual adversarial attack methods.

(2) We develop an interactive system called \texttt{HITL-GAT} based on the general approach to human-in-the-loop generation of adversarial texts.
This system is successfully applied to a case study on one lower-resourced language.

(3) We demonstrate the utility of \texttt{HITL-GAT} through a case study on Tibetan script, employing three customized adversarial text generation methods and establishing its first adversarial robustness benchmark, providing a valuable reference for other lower-resourced languages.

(4) We open-source both the system and the case study under GNU General Public License v3.0 to facilitate future explorations.
Our code repository received 42 stars, and our 12 victim models were downloaded more than 5,000 times before paper submission on November 15, 2025.

\section{Related Work}

\subsection{Textual Adversarial Attack Frameworks}

TextAttack \citep{morris-etal-2020-textattack} and OpenAttack \citep{zeng-etal-2021-openattack} are two powerful and easy-to-use Python frameworks for textual adversarial attacks.
They are both for text classification, supporting English and Chinese, with similar toolkit functionality and complementary attack methods.
From a developer's perspective, TextAttack utilizes a relatively rigorous architecture to unify different attack methods, while OpenAttack is more flexible.
SeqAttack \citep{simoncini-spanakis-2021-seqattack} and RobustQA \citep{boreshban-etal-2023-robustqa} are textual adversarial attack frameworks for named entity recognition and question answering, respectively, supporting English only.
These frameworks provide an excellent platform to stress-test the adversarial robustness of models targeting higher-resourced languages.
However, the weaponization of lower-resourced languages against NLP security \citep{lent2025weaponizationnlpsecuritymedium,yoo-etal-2025-code,lent-etal-2025-nlp} highlights the urgent need for research in this area.
To our knowledge, \texttt{HITL-GAT} is the first interactive system to build adversarial robustness benchmarks from scratch for a truly low-resource language.

\subsection{Human-in-the-Loop Adversarial Text Generation}

\citet{wallace-etal-2019-trick} guide human authors to keep crafting adversarial questions to break the question answering models with the aid of visual model predictions and interpretations.
They conduct two rounds of adversarial writing.
In the first round, human authors attack a traditional ElasticSearch model \texttt{A} to construct the adversarial set \texttt{x}.
Then, they use \texttt{x} to evaluate \texttt{A}, a bidirectional recurrent neural network model \texttt{B}, and a deep averaging network model \texttt{C}.
In the second round, they train \texttt{A}, \texttt{B}, and \texttt{C} on a larger dataset.
Human authors attack \texttt{A} and \texttt{B} to construct the adversarial set \texttt{x} and \texttt{x'}.
Then, they use \texttt{x} and \texttt{x'} to evaluate \texttt{A}, \texttt{B}, and \texttt{C}.
We see their human-in-the-loop approach as an embryo of adversarial robustness benchmark evolution, despite the high labor cost of relying on human authors to think and write adversarial texts.
Most goals of using a human-in-the-loop approach in NLP tasks are to improve the model performance in various aspects \citep{wang-etal-2021-putting}.
With these goals, language models evolve.
As continuous advancement of model capabilities, it is imperative to explore the paradigm for benchmark evolution.
To our knowledge, even though our work is preliminary, we are the first to explore the evolution of adversarial robustness benchmarks.

\section{Implementation}
\label{implementation}

\paragraph{Definition}

Due to the general adaptability of language models to the text classification task, our work focuses on the adversarial robustness evaluation of language models on this task.
The definition of textual adversarial attacks on text classification is as follows.
For a text classifier $F$, let $x$ ($x\in{X}$, $X$ includes all possible input texts) be the original input text and $y$ ($y\in{Y}$, $Y$ includes all possible output labels) be the corresponding output label of $x$, denoted as $F(x)={\mathop{\arg\max}_{\dot{y}\in{Y}}{P(\dot{y}|x)}}={y}$.
For a successful textual adversarial attack, let $x'=x+\delta$ be the perturbed input text, where $\delta$ is the imperceptible perturbation, denoted as $F(x')={\mathop{\arg\max}_{\dot{y}\in{Y}}{P(\dot{y}|x')}}\neq{y}$.

\paragraph{Overview}

Our system for human-in-the-loop generation of adversarial texts, \texttt{HITL-GAT}, contains four stages in one pipeline: \textbf{victim model construction}, \textbf{adversarial example generation}, \textbf{high-quality benchmark construction}, and \textbf{adversarial robustness evaluation}.
Figure \ref{flowchart} depicts the flowchart of \texttt{HITL-GAT}.
These four stages will be detailed in the following four subsections respectively.
Our flexible interactive system allows users to either go through the entire pipeline or directly start at any stage.
Gradio \citep{abid2019gradiohasslefreesharingtesting} is an open-sourced Python package that allows developers to quickly build a web demo or application for machine learning.
LlamaBoard is the user-friendly GUI (Graphical User Interface) of LlamaFactory \citep{zheng-etal-2024-llamafactory}.
The GUI of our system is powered by Gradio and draws inspiration from the design of LlamaBoard.

\subsection{Construct Victim Models}

This stage aims at constructing victim language models via a fine-tuning paradigm.

When a new language model \texttt{B} emerges, in order to better evaluate the adversarial robustness of \texttt{B}, we need to quantitatively and thoroughly perform evaluation on multiple downstream tasks.
For the purpose of stress-testing the adversarial robustness of \texttt{B} more effectively, i.e., constructing a stronger adversarial robustness benchmark with high quality, we can choose at least one previous SOTA or similar-structured language model \texttt{A} to implement textual adversarial attacks on it to generate updated adversarial texts.
We can also follow this stage when a new downstream dataset \texttt{n} is available.

In this stage, we fine-tune \texttt{A} and \texttt{B} on the training set of the same downstream datasets \texttt{1,2,...,n} to construct victim language models.
The victim model construction stage is depicted in the first part of Figure \ref{flowchart}.

\subsection{Generate Adversarial Examples}

This stage aims at automatically generating the first-round adversarial texts with the help of various textual adversarial attack methods.

The way human authors keep thinking and writing adversarial texts \citep{wallace-etal-2019-trick} is high-labor-cost.
With the emergence of automated textual adversarial attacks, such as TextFooler \citep{DBLP:conf/aaai/JinJZS20}, BERT-ATTACK \citep{li-etal-2020-bert-attack}, SemAttack \citep{wang-etal-2022-semattack}, and TextCheater \citep{10345721}, adversarial text generation has become relatively easy.
We can directly enter this stage when a new textual adversarial attack \texttt{N} appears.

In this stage, we implement textual adversarial attacks \texttt{I,II,...,N} on the victim language models constructed from language model \texttt{A} upon the test set of downstream datasets \texttt{1,2,...,n} to generate the first-round adversarial texts automatically.
The adversarial example generation stage is depicted in the second part of Figure \ref{flowchart}.

\subsection{Construct High-Quality Benchmarks}

This stage aims at constructing a high-quality adversarial robustness benchmark by customizing filter conditions and conducting human annotation.

The construction process of AdvGLUE \citep{NEURIPS_DATASETS_AND_BENCHMARKS2021_335f5352}, a widely used adversarial robustness benchmark, tells us that most textual adversarial attack methods are prone to generating invalid or ambiguous adversarial texts, with around 90\% either changing the original semantics or hindering the annotators' unanimity.
Therefore, human annotation is indispensable and can make benchmarks more practical and relevant.
In order to reduce the cost of human annotation, the first-round adversarial texts need to be screened automatically first using appropriate filter conditions.
Due to the fact that humans perceive texts through their eyes and brains, both filter conditions and human annotation should follow the visual and semantic similarity between adversarial texts and original texts.
Filter conditions can be the following metrics: Edit Distance, Normalized Cross-Correlation Coefficient (from the perspective of visual similarity); Cosine Similarity, BERTScore \citep{DBLP:conf/iclr/ZhangKWWA20} (from the perspective of semantic similarity); and so on.
Human annotation still requires additional consideration of annotators' unanimity so that adversarial texts can be deemed human-acceptable.
For example, given an original text and an adversarial text, we ask several annotators to score the human acceptance of the adversarial text based on the visual and semantic similarity between the two texts, from 1 to 5.
The higher the score, the higher the human acceptance.
If all annotators score the human acceptance of the adversarial text as 4 or 5, the adversarial text will be included in the adversarial robustness benchmark.

In this stage, we screen out the examples that do not satisfy the customized filter conditions from the first-round adversarial texts, and then manually annotate the remaining examples to construct the high-quality adversarial robustness benchmark \texttt{x}.
The high-quality benchmark construction stage is depicted in the third part of Figure \ref{flowchart}.

\subsection{Evaluate Adversarial Robustness}

This stage aims at quantitatively and thoroughly evaluating the adversarial robustness of new language models using the constructed high-quality adversarial robustness benchmark.

The adversarial robustness benchmark \texttt{x} is a collection of $n$ subsets, each of which contains high-quality adversarial texts generated from the test set of the corresponding downstream dataset.
We take the average accuracy on $n$ subsets as the adversarial robustness ($AdvRobust$) of the new language model \texttt{B} on \texttt{x}, denoted as:
\begin{equation}
	\label{AdvRobust}
	AdvRobust=\frac{\sum_{~i=1}^{~n}{Accuracy_{~i}}}{n}.
\end{equation}

In this stage, we utilize the constructed high-quality adversarial robustness benchmark \texttt{x} to evaluate the adversarial robustness of the language model \texttt{B} quantitatively and thoroughly.
The adversarial robustness evaluation stage is depicted in the fourth part of Figure \ref{flowchart}.

\section{Case Study}

In this section, we go through the entire pipeline under the existing conditions to construct the first adversarial robustness benchmark for Tibetan script and conduct the adversarial robustness evaluation on Tibetan language models.
We will introduce the existing conditions and the whole process in the following two subsections respectively.

\subsection{Existing Conditions}

Below is the involved language models, downstream datasets, and attack methods.

\subsubsection{Language Models}

\begin{itemize}[label=,labelsep=0pt,itemsep=0pt,topsep=0pt]
	\item \textbf{\texttt{Tibetan-BERT}}\footnote{\url{https://huggingface.co/UTibetNLP/tibetan_bert}} \citep{10.1145/3548608.3559255}.
	A BERT-based monolingual model targeting Tibetan.
	It is the first Tibetan BERT model and achieves a good result on the specific downstream Tibetan text classification task.
	\item \textbf{\texttt{CINO}}\footnote{\url{https://huggingface.co/hfl/cino-small-v2}\\\url{https://huggingface.co/hfl/cino-base-v2}\\\url{https://huggingface.co/hfl/cino-large-v2}} \citep{yang-etal-2022-cino}.
	A series of XLM-RoBERTa-based multilingual models including Tibetan.
	It is the first multilingual model for Chinese minority languages and achieves a SOTA performance on multiple downstream monolingual or multilingual text classification task.
\end{itemize}

\subsubsection{Downstream Datasets}

\begin{itemize}[label=,labelsep=0pt,itemsep=0pt,topsep=0pt]
	\item \textbf{\texttt{TNCC-title}}\footnote{\url{https://github.com/FudanNLP/Tibetan-Classification}} \citep{DBLP:conf/cncl/QunLQH17}.
	A Tibetan news title classification dataset.
	This dataset contains a total of 9,276 Tibetan news titles, which are divided into 12 classes.
	\item \textbf{\texttt{TU\_SA}}\footnote{\url{https://github.com/UTibetNLP/TU_SA}} \citep{MESS202302007}.
	A Tibetan sentiment analysis dataset.
	It is built by translating and proofreading 10,000 sentences from two public Chinese sentiment analysis datasets.
	In this dataset, negative or positive class each accounts for 50\%.
\end{itemize}

\subsubsection{Attack Methods}

Over the past few years, we have developed several Tibetan textual adversarial attack methods, aiming to draw attention to the NLP security in lower-resourced languages, as listed below.
Our past work \citep{cao-etal-2023-pay-attention} is the only one engaged with a truly low-resource language among the research samples in the literature \textit{NLP Security and Ethics, in the Wild} (\citeauthor{lent-etal-2025-nlp}, TACL 2025, page 719).

\begin{itemize}[label=,labelsep=0pt,itemsep=0pt,topsep=0pt]
	\item \textbf{\texttt{TSAttacker}} \citep{cao-etal-2023-pay-attention}.
	An embedding-similarity-based Tibetan textual adversarial attack.
	It utilizes the cosine distance between static syllable embeddings to generate substitution syllables.
	\item \textbf{\texttt{TSTricker}} \citep{10.1145/3589335.3652503}.
	A context-aware-based Tibetan textual adversarial attack.
	It utilizes two BERT-based masked language models with tokenizers of two different granularities to generate substitution syllables or words respectively.
	\item \textbf{\texttt{TSCheater}} \citep{10889732}.
	A visual-similarity-based Tibetan textual adversarial attack.
	It utilizes a self-constructed Tibetan syllable visual similarity database to generate substitution candidates.
\end{itemize}

\subsection{Whole Process}

\begin{figure}[h]
	\centering
	\includegraphics[width=\linewidth]{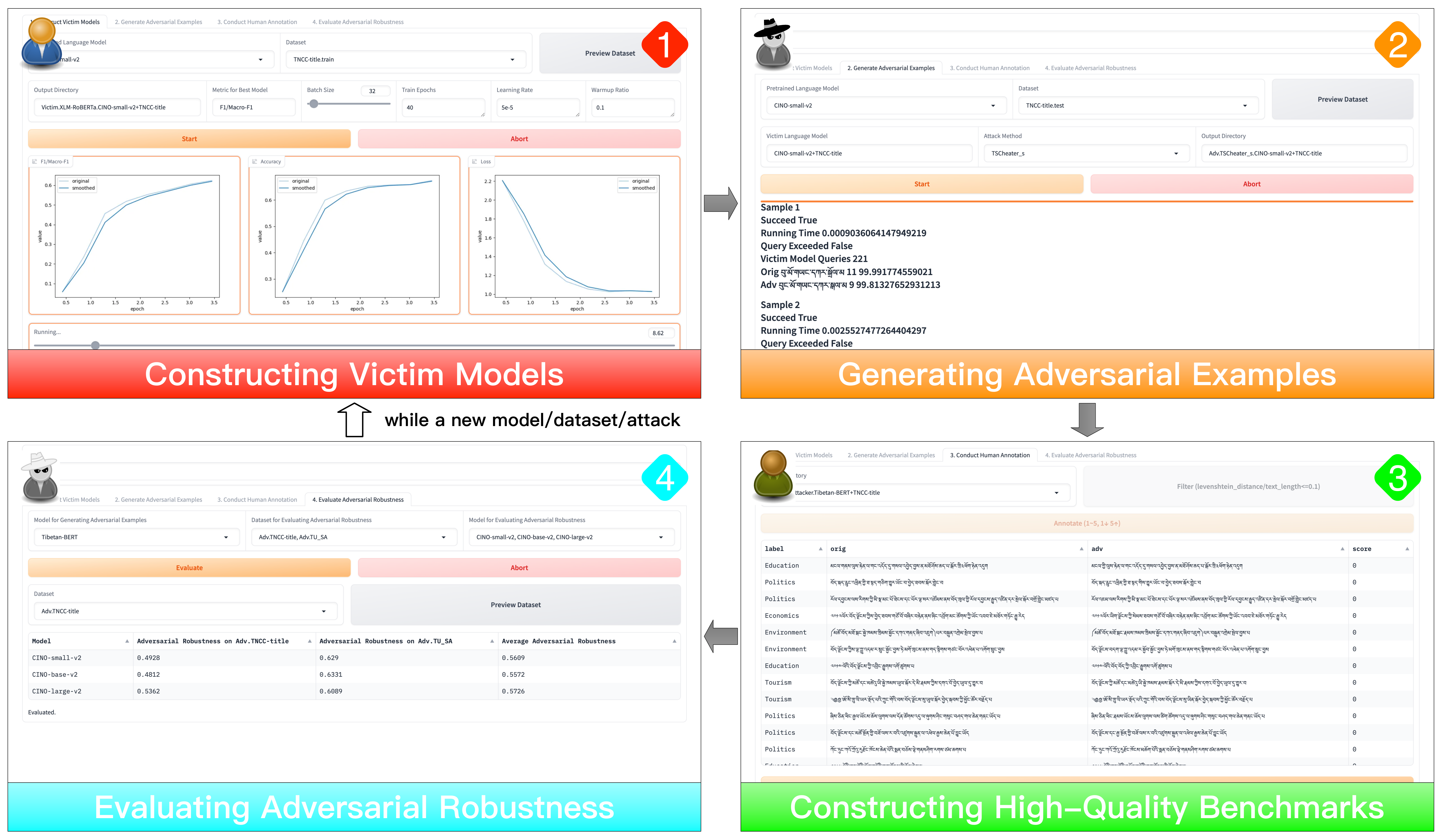}
	\caption{Screenshots of \texttt{HITL-GAT}.}
	\label{screenshots}
\end{figure}

Figure \ref{flowchart} and Section \ref{implementation} introduce the four stages of \texttt{HITL-GAT}.
Below we use a case study on Tibetan script to illustrate the whole process, which is also demonstrated in the video and Figure \ref{screenshots}.

In the victim model construction stage, we choose the language model and downstream dataset, and then the default fine-tuning hyperparameters will be loaded.
Once the ``Start'' button is clicked, the fine-tuning starts and the GUI displays a progress bar, metric plots (F1/macro-F1, Accuracy, and Loss) and running logs.
Here, we fine-tune \texttt{Tibetan-BERT} and \texttt{CINO} series on the training set of \texttt{TNCC-title} and \texttt{TU\_SA} to construct the victim language models.

Next, in the adversarial example generation stage, we choose the language model and downstream dataset, and then the victim language model will be loaded.
Once the ``Start'' button is clicked, the attack starts and the GUI displays generated examples.
Here, we implement \texttt{TSAttacker}, \texttt{TSTricker}, and \texttt{TSCheater} on the victim language models constructed from \texttt{Tibetan-BERT} upon the test set of \texttt{TNCC-title} and \texttt{TU\_SA} to generate the first-round adversarial texts.

Thereafter, in the high-quality benchmark construction stage, we screen out the examples that do not satisfy the customized filter condition $levenshtein\_distance/text\_length<=0.1$ from the first-round adversarial texts, and then manually annotate the remaining examples to construct the first Tibetan adversarial robustness benchmark called \texttt{AdvTS}.
Given an original text and an adversarial text, we ask 3 annotators to score the human acceptance of the adversarial text based on the visual and semantic similarity between the two texts, from 1 to 5.
The higher the score, the higher the human acceptance.
If all annotators score the human acceptance of the adversarial text as 4 or 5, the adversarial text will be included in \texttt{AdvTS}.
Below is the guidelines for human annotation.

\begin{itemize}[label=,labelsep=0pt,itemsep=0pt,topsep=0pt]
	\item
	\textbf{Score 1: Definite Reject.}
	Humans can intuitively perceive that the perturbations significantly alter the appearance or semantics of the original text.
	\item
	\textbf{Score 2: Reject.}
	Humans can intuitively perceive that the perturbations do alter the appearance or semantics of the original text.
	\item
	\textbf{Score 3: Marginal Reject or Accept.}
	Humans can intuitively perceive that the perturbations alter the appearance or semantics of the original text not too much.
	\item
	\textbf{Score 4: Accept.}
	After careful observation or thought for 5 seconds, humans find that perturbations only slightly alter the appearance or semantics of the original text.
	\item
	\textbf{Score 5: Definite Accept.}
	After careful observation for 5 seconds, humans can not find that perturbations alter the appearance of the original text.
	Or, after careful thought for 5 seconds, humans find that perturbations do not alter the semantics of the original text.
\end{itemize}

Finally, in the adversarial robustness evaluation stage, we utilize \texttt{AdvTS} to evaluate the adversarial robustness of \texttt{CINO} series with Equation \ref{AdvRobust}.
The $AdvRobust$ of \texttt{CINO-small-v2}, \texttt{CINO-base-v2}, and \texttt{CINO-large-v2} is 0.5609, 0.5572, and 0.5726 respectively.

While a new language model, downstream dataset, or textual adversarial attack method emerges, we can enter the loop again to make the adversarial robustness benchmark evolve.

\section{Discussion}

Due to the fact that humans perceive texts through their eyes and brains, when the perturbed text tends to the original text in visual or semantic similarity, we consider such perturbations to be imperceptible.
To construct imperceptible perturbations, we can start from the following three aspects.

\begin{itemize}[label=,labelsep=0pt,itemsep=0pt,topsep=0pt]
	
	\item
	\textbf{Transplanting existing general methods.}
	From the perspective of semantic approximation, using synonyms for substitution is a general method.
	Sources of synonyms can be static word embeddings \citep{alzantot-etal-2018-generating}, dictionaries \citep{ren-etal-2019-generating}, and predictions of masked language models \citep{li-etal-2020-bert-attack}.
	\item
	\textbf{Using intrinsic textual features.}
	Different languages have different features inherent in their texts.
	For example, in abugidas (Tibetan script, Devanagari script, etc.), many pairs of confusable letters result in visually similar syllables \citep{kaing-etal-2024-robust,10889732}.
	
	\item
	\textbf{Using extrinsic encoding features.}
	In the process of historical development, there are many cases of ``same language with different encodings''.
	For example, due to the technical problems in history, there are two Tibetan coded character sets in national standards of P.R.C (basic set: GB 16959-1997 and extension set: GB/T 20542-2006, GB/T 22238-2008);
	due to the simplification of Chinese characters, simplified and traditional Chinese exist.
	Encoding issues between different languages also deserve attention.
	For example, the Latin letter \texttt{x} (U+0078) and the Cyrillic letter \texttt{x} (U+0445) look the same; ZWNJ (zero width non-joiner, U+200C) is used extensively for certain prefixes, suffixes and compound words in Persian, but it is invisible and useless in most other languages.
	
\end{itemize}

\section{Conclusion}

This paper introduces \texttt{HITL-GAT}, an interactive system for human-in-the-loop generation of adversarial texts.
Our approach employs a four-stage iterative loop: victim model construction, adversarial example generation, high-quality benchmark construction, and adversarial robustness evaluation.
The loop ensures adversarial robustness benchmarks to co-evolve with advancements in language models, downstream datasets, and textual adversarial attack methods.
Additionally, we demonstrate the utility of \texttt{HITL-GAT} through a case study on Tibetan script, employing three customized adversarial text generation methods and establishing its first adversarial robustness benchmark.
Our work provides a valuable reference for other lower-resourced languages, especially languages in the Asia-Pacific that use abugidas as their writing system.
The weaponization of lower-resourced languages against NLP security highlights the critical gap and the urgent need for research in this area.

\newpage

\section*{Limitations}

The system and the case study presented in this paper are the crystallization of our research on the adversarial robustness of Tibetan language models over the past few years.
The summarized approach is only applicable to classification tasks.
Given the heightened sensitivity necessary for working with lower-resourced languages, our case study is conducted on insensitive tasks with ethical best practices in mind.
Due to the existing conditions of lower-resourced languages, our case study can only be conducted this far.
However, this does not prevent it from serving as an early paradigm for researching the evolution of adversarial robustness benchmarks.
We will continue to develop \texttt{HITL-GAT} and conduct more case studies on other minority languages.

\section*{Ethical Considerations}

Our work adheres to the ACM Code of Ethics.
The purpose of this paper is to promote research on NLP security, especially for lower-resourced languages.
The textual adversarial attack methods mentioned in this paper must be used positively, thus preventing any malicious misuse.
Additionally, adherence to the model or dataset license is mandatory when using our system or fork versions, thus preventing any potential misuse.

\section*{Acknowledgments}

This work benefited greatly from the encouragement, advice, and help of others.
Thank you to Heather Lent at Aalborg University for her encouragement and detailed feedback on our manuscript.
Thank you to Chen Zhang at Peking University and Andong Chen at Harbin Institute of Technology for their early advice.
We also thank the anonymous reviewers of *ACL for their insightful comments.

Finally, thanks to the following open-sourced projects: OpenAttack \citep{zeng-etal-2021-openattack}, Gradio \citep{abid2019gradiohasslefreesharingtesting}, LlamaFactory \citep{zheng-etal-2024-llamafactory}, Transformers \citep{wolf-etal-2020-transformers}, Datasets \citep{lhoest-etal-2021-datasets}, etc.

This work is supported by the Science and Technology Strategic Consulting Project of the Chinese Academy of Engineering (2025-XZ-16-06), the Key Project of Xizang Natural Science Foundation (XZ202401ZR0040), the National Social Science Foundation of China (22\&ZD035), the National Natural Science Foundation of China (61972436), and the MUC (Minzu University of China) Foundation (2025XYCM39).



\bibliography{custom}

\appendix

\end{document}